\documentclass[letterpaper, 10 pt, conference]{ieeeconf}  

\IEEEoverridecommandlockouts                              

\usepackage{graphics} 
\usepackage{epsfig} 
\usepackage{mathptmx} 
\usepackage{times} 
\usepackage{microtype}
\usepackage{algorithmic}
\usepackage{textcomp}
\usepackage{xcolor}
\usepackage{gensymb}
\usepackage{caption}
\usepackage{hyperref}
\usepackage{mwe}
\usepackage{verbatim}
\usepackage{amsbsy}
\usepackage{siunitx}
\usepackage{tabularx,booktabs}
\usepackage{dblfloatfix}
\usepackage{cite}
\usepackage{bm}
\usepackage{amsmath,amssymb,amsfonts}
\usepackage{subcaption}
\usepackage{float}

\usepackage{algorithm, algorithmic}
\usepackage[autolanguage]{numprint}

\usepackage{spreadtab}
\usepackage{multirow}
\usepackage{cite}
\usepackage{hyperref}
\usepackage{xcolor}
\usepackage{algorithmic}
\usepackage{textcomp}
\usepackage{xcolor}
\usepackage{gensymb}
\usepackage{caption}
\usepackage{mwe}
\usepackage{amsfonts}
\usepackage{amssymb}
\usepackage{verbatim}
\usepackage{amsbsy}
\usepackage{siunitx}
\usepackage{tabularx, booktabs}
\usepackage{soul}
\usepackage{tikz}
\usetikzlibrary{calc}

\makeatletter
\newif\if@anonymize

\@anonymizefalse  

\if@anonymize
  \newcommand{\highlight@DoHighlight}{
    \fill [outer sep = -15pt, inner sep = 0pt, color=black]
          ($(begin highlight)+(0,8pt)$) rectangle ($(end highlight)+(0,-3pt)$) ;
  }

  \newcommand{\highlight@BeginHighlight}{
    \coordinate (begin highlight) at (0,0) ;
  }

  \newcommand{\highlight@EndHighlight}{
    \coordinate (end highlight) at (0,0) ;
  }

  \newdimen\highlight@previous
  \newdimen\highlight@current
  \newlength{\item@width}

  \DeclareRobustCommand*\anonymize{%
    \SOUL@setup
    \def\SOUL@preamble{%
      \begin{tikzpicture}[overlay, remember picture]
        \highlight@BeginHighlight
        \highlight@EndHighlight
      \end{tikzpicture}%
    }%
    \def\SOUL@postamble{%
      \begin{tikzpicture}[overlay, remember picture]
        \highlight@EndHighlight
        \highlight@DoHighlight
      \end{tikzpicture}%
    }%
    \def\SOUL@everyhyphen{%
      \discretionary{%
        \SOUL@setkern\SOUL@hyphkern
        \SOUL@sethyphenchar
        \tikz[overlay, remember picture] \highlight@EndHighlight ;%
      }{%
      }{%
        \SOUL@setkern\SOUL@charkern
      }%
    }%
    \def\SOUL@everyexhyphen##1{%
      \SOUL@setkern\SOUL@hyphkern
      \settowidth{\item@width}{##1}%
      \makebox[\item@width]{}%
      \discretionary{%
        \tikz[overlay, remember picture] \highlight@EndHighlight ;%
      }{%
      }{%
        \SOUL@setkern\SOUL@charkern
      }%
    }%
    \def\SOUL@everysyllable{%
      \begin{tikzpicture}[overlay, remember picture]
        \path let \p0 = (begin highlight), \p1 = (0,0) in \pgfextra
          \global\highlight@previous=\y0
          \global\highlight@current =\y1
        \endpgfextra (0,0) ;
        \ifdim\highlight@current < \highlight@previous
          \highlight@DoHighlight
          \highlight@BeginHighlight
        \fi
      \end{tikzpicture}%
      \settowidth{\item@width}{\the\SOUL@syllable}%
      \makebox[\item@width]{}%
      \tikz[overlay, remember picture] \highlight@EndHighlight ;%
    }%
    \SOUL@
  }
\else
  \newcommand{\anonymize}[1]{#1}
\fi
\makeatother 

\title{\LARGE \bf
Virtual Reality Platform to Develop and Test\\Applications on Human-Robot Social Interaction
}

\author{\anonymize{Jair A. Bottega$^{1}$, Raul Steinmetz$^{1}$, Alisson H. Kolling$^{1}$, Victor A. Kich$^{1}$, }\\\anonymize{Junior C. de Jesus$^{2}$, Ricardo B. Grando$^{3}$, Daniel F. T. Gamarra$^{1}$}
\thanks{\anonymize{$^{1}$Jair A. Bottega, Raul Steinmetz, Victor A. Kich, Alisson H. Kolling and Daniel F. T. Gamarra are with Federal University of Santa Maria - UFSM, Santa Maria, RS, Brazil.}
        {\tt\small \anonymize{jairaugustobottega@gmail.com}}}%
\thanks{$^{2}$\anonymize{Junior C. de Jesus is with Universidade Federal do Rio Grande - FURG, RS, Brazil.}
        {\tt\small \anonymize{dranaju@gmail.com}}}%
\thanks{$^{3}$\anonymize{Ricardo B. Grando is with Technological University of Uruguay - UTEC, Rivera, Uruguay.}
        {\tt\small \anonymize{ricardo.bedin@utec.edu.uy}}}%
}

\begin{document}

\maketitle
\thispagestyle{empty}
\pagestyle{empty}

\begin{abstract}


Robotics simulation has been an integral part of research and development in the robotics area. The simulation eliminates the possibility of harm to sensors, motors, and the physical structure of a real robot by enabling robotics application testing to be carried out quickly and affordably without being subjected to mechanical or electronic errors. Simulation through virtual reality (VR) offers a more immersive experience by providing better visual cues of environments, making it an appealing alternative for interacting with simulated robots. 
This immersion is crucial, particularly when discussing sociable robots, a subarea of the human-robot interaction (HRI) field.
The widespread use of robots in daily life depends on HRI. In the future, robots will be able to interact effectively with people to perform a variety of tasks in human civilization. It is crucial to develop simple and understandable interfaces for robots as they begin to proliferate in the personal workspace.
Due to this, in this study, we implement a VR robotic framework with ready-to-use tools and packages to enhance research and application development in social HRI. Since the entire VR interface is an open-source project, the tests can be conducted in an immersive environment without needing a physical robot.

\end{abstract}

\section*{SUPPLEMENTARY MATERIAL}

Released code and Docker image is available at \anonymize{https://github.com/jajaguto/jubileo}. 

\section{INTRODUCTION}\label{introduction}

Since its inception, simulation in the field of robotics has played a crucial role in research and development. For instance, to provide a secure and affordable human-robot interaction experience, robotic surgical training, remote manipulation interfaces, manufacturing training exercises, and human-robot interaction research are frequently done on a screen in a virtual environment. Virtual reality (VR), however, offers a more immersive experience by better simulating real situations, making it a desirable choice for engaging with virtual robots. Prior research studies illustrate that Human-Robot Interaction (HRI) can be enhanced with VR interfaces~\cite{tan2012sigverse,liu2017understanding}.

Through simulation, robotic applications can be tested rapidly and affordably without being subjected to mechanical or electronic mistakes, eliminating the possibility of a real robot's sensors, motors, or physical structure being damaged. The importance of this aspect can be seen in~\cite{MALIK2021102092}. Also, the developers have no need to worry about human safety, as discussed in~\cite{ZACHARAKI2020104667}.
Many HRI applications involve users interacting with simulated robots.
HRI is defined by Goodrich and Schultz~\cite{goodrich} as a branch of study devoted to comprehending, developing, and evaluating robotic systems for usage with humans. Communication between people and robots is necessary for human-robot social interaction, and it can be done in a variety of ways. Humans use their voice, gestures, and body language to convey their sentiments and attitudes to others and to decipher the emotions, intentions, and desires of other individuals~\cite{plutchik1984emotions}. Humanoid robots were created in order to enable the direct social connection between a human and a robot. These robots have human-like abilities and resemble humans physically

\begin{figure}[tbp!]
    \centering
    \includegraphics[width=\linewidth]{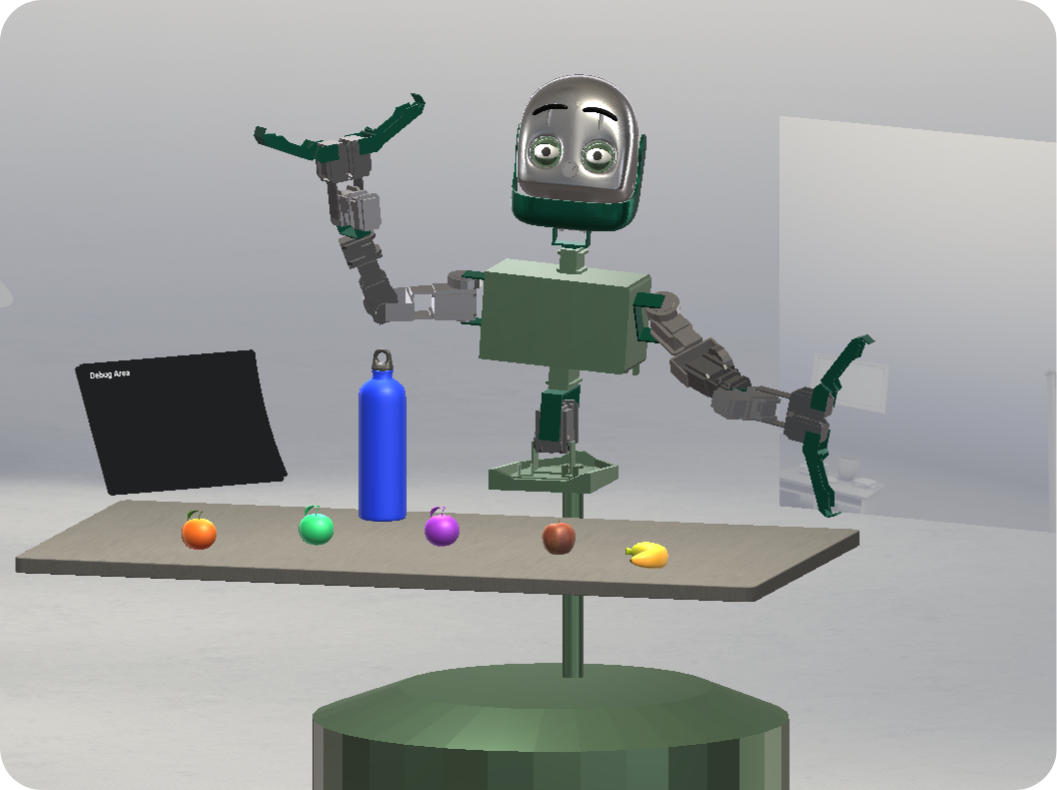}
    \caption{Social robot in VR simulated environment.}
    \label{fig:structure}
    \vspace{-7mm}
\end{figure}

Thus, this work presents an open-source VR interface that implements the model of a social robot with an animatronic face and a torso with two manipulators as arms.
The developed platform enables researchers to create and test applications in the field of HRI in a realistic manner, simulating actual robot interaction.
That strategy enhances the evaluation of social robotics systems because it increases immersion and enables an interaction to feel more like it would be in a real-world setting.

The simulated robot includes a complete operational system with ready-to-use tools that facilitates the development of social HRI applications.
Its system includes, for example, computational vision capabilities, voice recognition, speech synthesis, visual tracking control, joint control, and facial expression settings. Also, the robot face was designed with the intent to escape from the Uncanny Valley~\cite{mori2012uncanny} concept, which is a state where an object's appearance is very similar to a human's, resulting in a negative emotional response to it. Thus, it has a friendly appearance that aims to increase comfort and encourage humans to engage in a conversation so that the robot can respond socially, processing and returning an answer by voice, gesture, or facial expressions.

The work is structured as follows: the related works are discussed in Section~\ref{related_works}.
In the Section~\ref{methodology} 
are explained the systems and techniques used to build the complete VR framework. The experiments realized are described in Section~\ref{experiments}, 
and a brief discussion about the experiments is made in Section~\ref{discussion}.
Finally, the main contributions and future works are summarized in Section~\ref{conclusions}.        

\section{RELATED WORKS}\label{related_works}


The emerging technologies of Augmented Reality (AR), Virtual Reality (VR), and Mixed Reality (XR) are making possible the study of fields in robotics without the need for expensive hardware. One of those areas of research is the study of human-robot interactions (HRI). The work~\cite{hri_vr_tools_study} evaluates existing VR-based solutions for simulating approaches that target or embody HRI. Beyond just interaction, the collaboration between humans and robots is significant for developing promising robots. Dianatfar \textit{et al.}~\cite{review_vr_hri_solutions} reviews the current status of virtual and augmented reality solutions in human-robot collaboration.

In~\cite{qualitative_study_colab}, Arntz \textit{et al.} explores the interactions of nine people with an AI-based representation of a robotic arm in a virtual reality environment as they jointly assemble a product with their robotic partner. This work demonstrates the possibilities of VR when studying HRI, giving meaning to the participants' experiences without needing a real robot. The work of Etzi \textit{et al.}~\cite{test_hri} assesses the human-robot interactions during the execution of a collaborative task and demonstrates that humans performance varies with the kinematic aspects of a robot's motion, but physiological responses don't alter. With such results in mind, we can assume that a VR simulation of a robot has a semblance of interaction with a real robot.

With the increasing presence of robots in our daily lives, their social aspects need to be worked on. Henschel \textit{et al.} works\cite{HENSCHEL2020373}, studying human social cognition when interacting with robots to understand how best to introduce robots to complex social settings. One way of facilitating the research on the social interactions between humans and robots is the employment of VR settings. A Social Virtual Reality Robots (V2R)\cite{v2r_autism} was conceived by Shahab \textit{et al.} with the ability to teach music to children with autism as well as perform an automatic assessment of their behaviors. Shariaty \textit{et al.} transferred the model of the Arash robot to a VR setting \cite{vr_arash}. This work found that the VR robot's acceptance is reasonably compatible with the actual robot since the performance of the VR robot did not have significant differences from the version of the real Arash robot. These two works prove the concepts presented in this work, that a social robot can interact with humans through a virtual reality setting and still maintain the social aspects of the real robot, enabling the studies of HRI through VR.

To be able to have meaningful interactions with a robot in a VR setting, new tools need to be developed. One such tool was designed by Inamura \textit{et al.} SIGVerse~\cite{japa_vr}, a cloud-based VR platform to research multimodal HRI. Developed with ROS and Unity, they show its feasibility in a robot competition setting. Another simulator produced for virtual reality is the work of Murnane \textit{et al.}~\cite{simulator_hri_vr}, which aims to collect training data for real-world robots. Our work differs from those previous works by focusing on the social aspect of HRI, exploring it through spoken language and the interaction with objects.

\section{METHODOLOGY}\label{methodology}

In this section, the concept of our approach is introduced, covering the development details of the research platform.
We present the robot model used in the framework, the developed system that operates the robot, and the simulated virtual environment created.

\subsection{Robot Model}

The robot model employed in this work contains an animatronic face and a torso with arms attached to a fixed base. 
The face used comes from the robot Doris~\cite{butia2020}, which participates in international robotics competitions in the @home category. The humanoid's animatronic face has complete control of the movements of the jaw, neck, eyes, eyelids, and eyebrows, making it possible to create different facial expressions and communicate with humans through fluid movements.

\begin{figure}[tp!]
    \centering
    \begin{subfigure}[t]{0.22\textwidth}
         \centering
         \includegraphics[width=\textwidth]{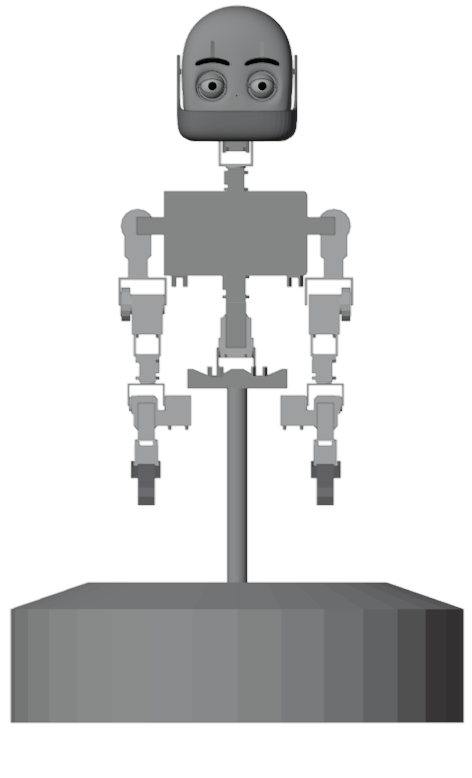}
         \caption{Front view.}
         \label{fig:front_view}
     \end{subfigure}
     \begin{subfigure}[t]{0.22\textwidth}
         \centering
         \includegraphics[width=\textwidth]{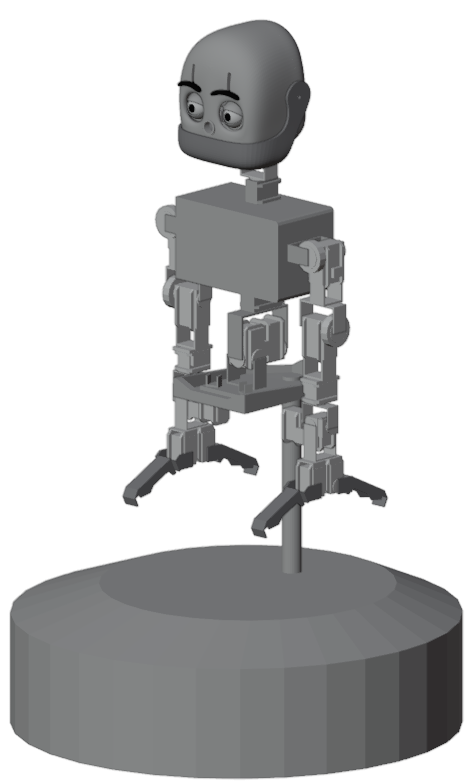}
         \caption{Isometric view.}
         \label{fig:simulated_face}
     \end{subfigure}
     \vspace{-1mm}
    \caption{3D model of the VR platform's robot.}
    \label{fig:3DModel}
    \vspace{-6mm}
\end{figure}

The torso and manipulators contained in our platform come from the Dimitri robot. Dimitri is an open-source full-body biped humanoid robot~\cite{dimitri2017}.
As our approach does not involve walking applications, the robot's legs were replaced with a fixed base, where the torso was attached.
Fig.~\ref{fig:3DModel} presents the resulting robot's 3d model applied to the VR platform.

Since it has a functional body and face, the resultant robot model allows a variety of HRI applications, since it is capable of doing facial expressions, visual tracking, speech movement, and manipulation tasks.
All our project is open-source, and the complete robot's URDF model is provided so that anyone can use them to create a simulated environment in any supported URDF robotics simulation platform.

\subsection{Robot Operating System}

Software for robots can be created using the Robot Operating System (ROS)~\cite{hart2015affordance}, which offers a flexible framework. An operational system's standard services, including hardware abstraction, low-level device control, messaging between processes, and package management, are provided by the ROS set of tools and libraries.
The set of ROS operations can be represented by a graphs architecture where the processing is performed on nodes that receive and send messages with data related to sensors, control, state, planning, actuator, and other topics that link those nodes.
Although it is feasible to connect ROS with real-time code, ROS is not a real-time operating system, despite the importance of reduced latency for robot control. To overcome this lack of a real-time system, ROS 2~\cite{kay2015real}, which is employed in this work, is currently being developed.

\begin{figure}[tbp!]
    \centering
    \includegraphics[width=\linewidth]{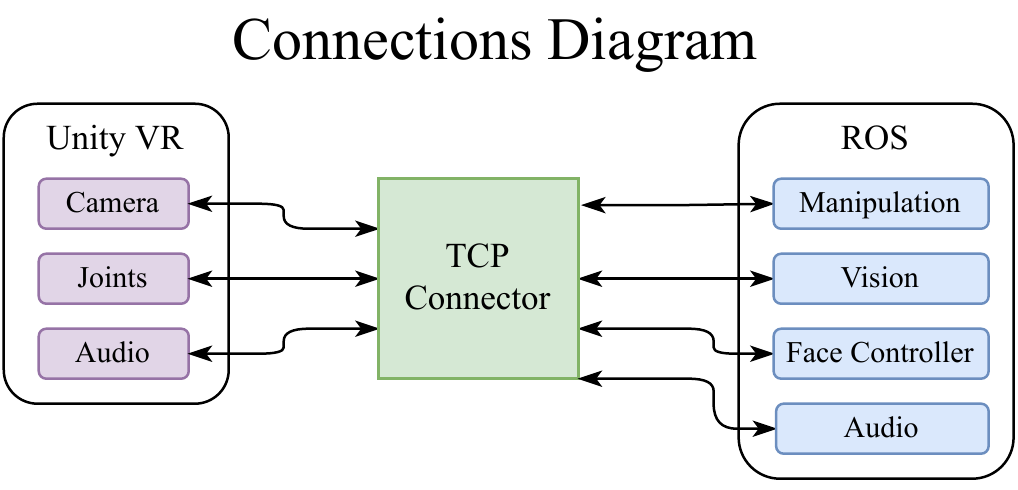}
    \vspace{-4mm}
    \caption{Scheme of ROS node connections structure.}
    \label{fig:system_diagram}
    \vspace{-6mm}
\end{figure}

The ROS packages developed for simulated robots provide researchers with a complete base of algorithms that enriches the robot's capabilities, such as computer vision, speech recognition, and robot joint control.
The Robot Operating System is capable of communicating with the VR software interface, sending and receiving data, allowing the system to control the simulated robot and process sensors information in real-time.

\subsection{Virtual Reality Environment}

The VR environment must contain a functional simulated robot with controllable joints, working sensors, physics, and collisions compatible with reality. Furthermore, to carry out immersive communication with the simulated robot, we need to build a virtual environment that contains elements that enable social interaction, such as 3D visual models of different objects and 3D avatars of people.

Unity is a multi-platform game and simulator creation engine. Since it supports integration with robotics, offering a variety of tools for developing VR aplications~\cite{nguven2017} and physics simulation, it was chosen for developing our simulation framework. Within the robotics tools offered by Unity, there is communication with ROS and the import of Unified Robotics Description Format (URDF) files~\cite{hussein2018}.

The TCP/IP protocol~\cite{meng2015}, a widely used method that establishes a connection between client and server before the data is delivered, is used for communication between ROS and Unity. So, the scripts in Unity can exchange information with the nodes from ROS, aiming to control the robot within the simulation.
A simplified overview of the ROS connection with the VR framework is present in Fig.~\ref{fig:system_diagram}. 

\begin{figure}[tbp!]
    \centering
    \includegraphics[width=0.91\linewidth]{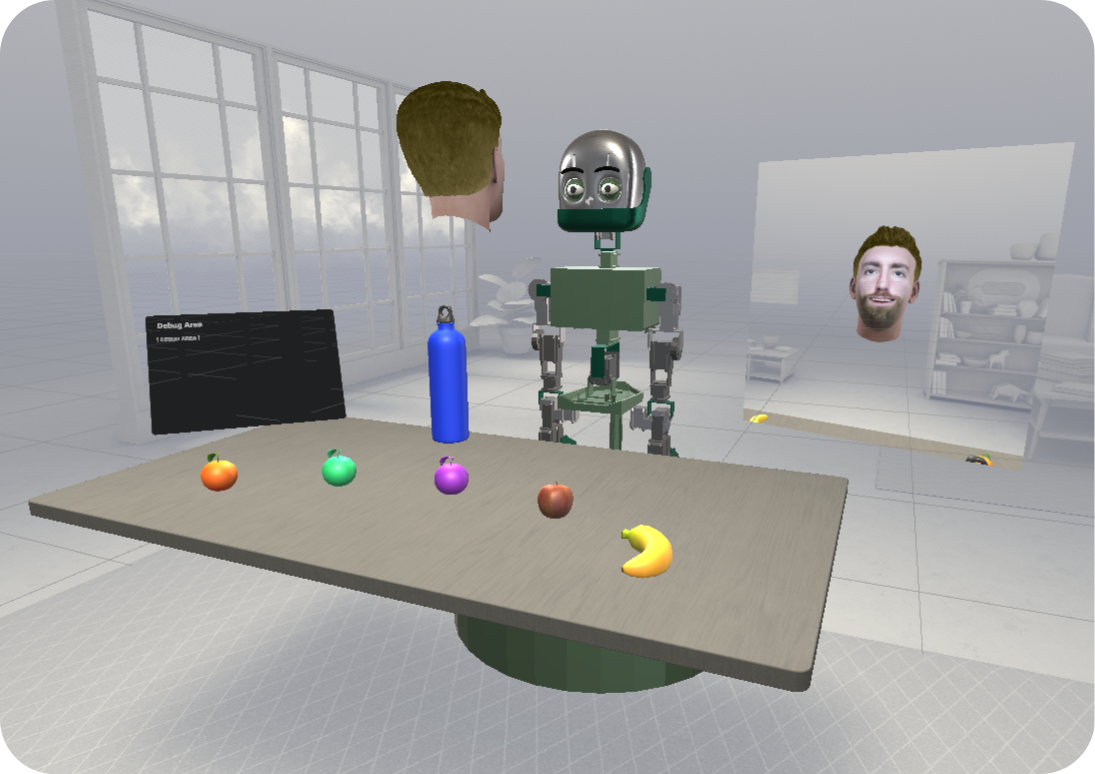}
    \caption{VR environment.}
    \label{fig:vr_environment}
    \vspace{-7mm}
\end{figure}

The virtual environment was built, containing a table with various objects that the robot can grasp or the user can grab in order to interact with the robot. Also, the user has a 3D avatar attached to his head inside the simulation, with a set of different facial expressions so that the robot can look directly at the user and communicate.
Next to the robot is a screen that displays in real-time the image that is captured by the camera sensor. And, in order for the researcher, for example, to check if the communication is working or to be able to detect errors in some developing ROS node, a debug console screen was placed inside the virtual environment, above the table. The VR environment can be seen in Fig.~\ref{fig:vr_environment}.

\section{EXPERIMENTS}\label{experiments}

In this section, some experiments are presented and could demonstrate our robot's social capabilities by having it interact with humans, do facial communication, and execute other activities related to the HRI field, as explored in famous studies~\cite{miseikis2020,sheridan}. 

\subsection{Visual Object and Face Tracking}\label{visual_tracking}
HRI's social applications highly depend on visual tracking algorithms, as shown in~\cite{zhou2018}, since they enable the robot to look at people and things. Our framework provides tracking options that can be made through neck and eye movement. The camera sensor in the robot's nose hole detects the target to be followed, and joints are controlled in accordance with the target location.

\begin{figure}[tp!]
    \centering
    \includegraphics[width=\linewidth]{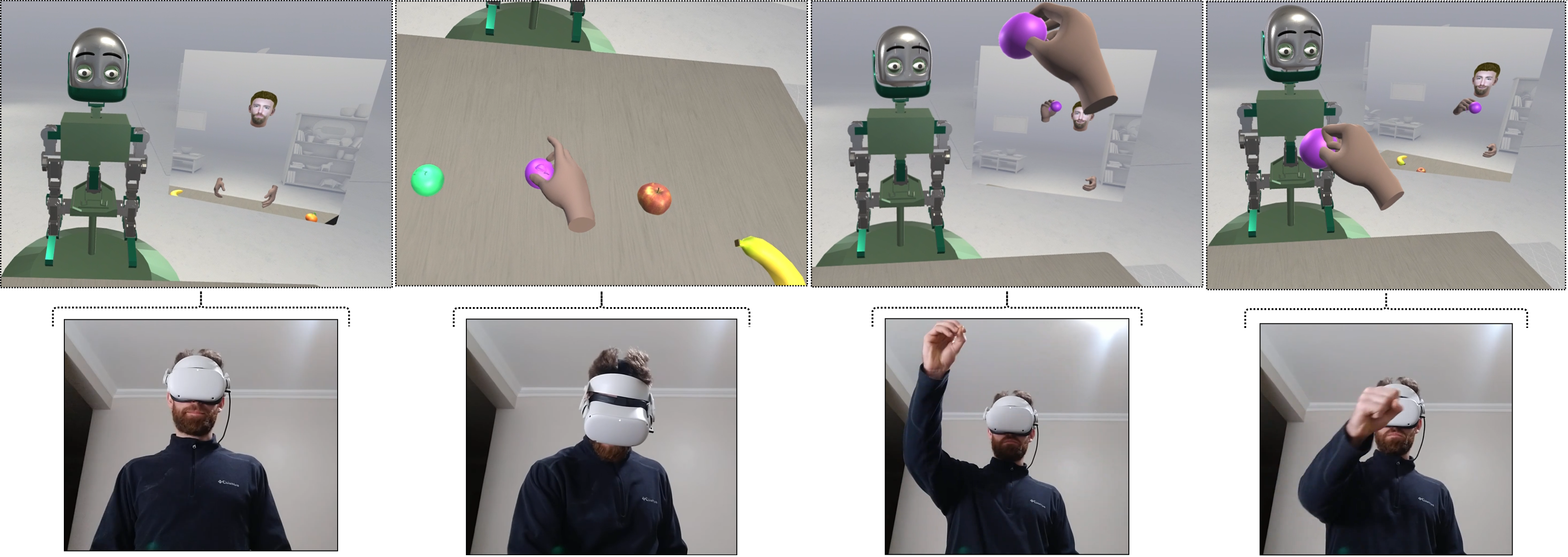}
    \caption{Visual object tracking while human manipulates it.}
    \label{fig:interaction}
    \vspace{-7mm}
\end{figure}

The tests in this section explore the robot's visual tracking effectiveness in the simulated world. The robot was initially instructed to focus on a specific color. A computer vision algorithm finds the item with the required color once the speech instruction is recognized. For instance, a human can test the algorithm by engaging with the virtual environment and manipulating the target objects, changing their positions. Fig.~\ref{fig:interaction} shows that interaction. The VR environment also includes a screen that displays the robot's camera perspective for improved user visualization.

Second, face detection and tracking is used to test the humanoid ability to track faces visually. By giving the "Look at me" voice command, the robot instantly starts gazing at the human in the simulation. For HRI, that defines an essential ability, as discussed previously.

\subsection{Face Recognition}\label{face_recognition}
The framework includes an easy-to-use, full-featured facial recognition capability to enhance user immersion when interacting with the robot. Thanks to this utility, researchers can create better interaction algorithms once the robot is able to save and use data about specific people.
For example, a routine can be developed where the robot can recognize familiar faces and address people by name, remembering personal details about them, and bring it to the human-robot dialogue, which is very important in the human view~\cite{khalifa2022face}.

\begin{figure}[bp!]
\vspace{-4mm}
    \centering
     \begin{subfigure}[t]{0.115\textwidth}
         \centering
         \includegraphics[width=\textwidth]{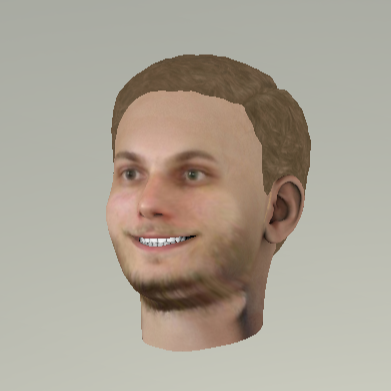}
         \label{fig:avatar1}
     \end{subfigure}
     \begin{subfigure}[t]{0.115\textwidth}
         \centering
         \includegraphics[width=\textwidth]{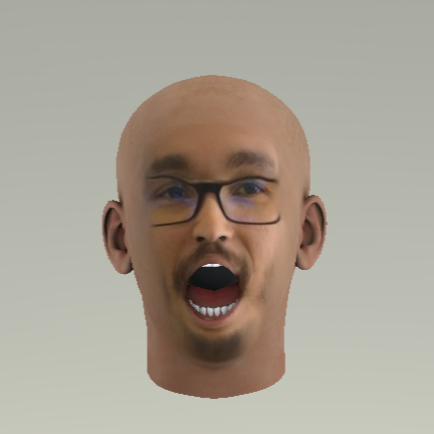}
         \label{fig:avatar2}
     \end{subfigure}
     \begin{subfigure}[t]{0.115\textwidth}
         \centering
         \includegraphics[width=\textwidth]{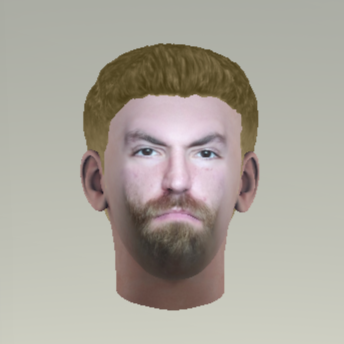}
         \label{fig:avatar3}
     \end{subfigure}
     \begin{subfigure}[t]{0.115\textwidth}
         \centering
         \includegraphics[width=\textwidth]{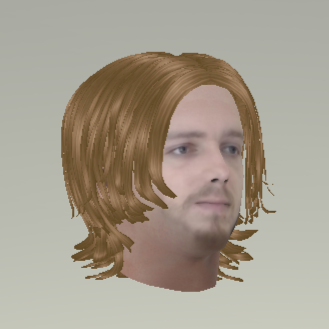}
         \label{fig:avatar4}
     \end{subfigure}
     \vspace{-5mm}
    \caption{Samples of the framework's 3D avatar set.}
    \label{fig:3DAvatars}
\end{figure}

Various 3D face models are available through the VR framework platform, including multiple people and facial expressions. These 3D avatars were used to explore the face recognition capabilities of the robot, and that presented good results, showing that it is capable of recognizing all the simulated avatars correctly. A sample of different 3D avatars provided by the VR platform is presented in Fig.~\ref{fig:3DAvatars}.

\subsection{Facial Expressions}\label{facial_expressions}
\begin{figure*}[tbp!]
    \centering
     \begin{subfigure}[t]{0.135\textwidth}
         \centering
         \includegraphics[width=\textwidth]{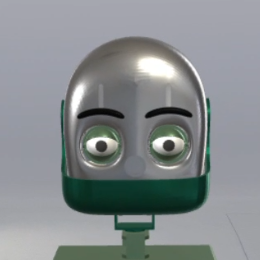}
         \caption{Neutral.}
         \label{fig:vr_face_neutral}
     \end{subfigure}
     \begin{subfigure}[t]{0.135\textwidth}
         \centering
         \includegraphics[width=\textwidth]{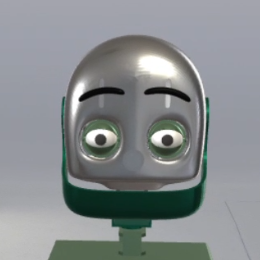}
         \caption{Joy.}
         \label{fig:vr_face_joy}
     \end{subfigure}
     \begin{subfigure}[t]{0.135\textwidth}
         \centering
         \includegraphics[width=\textwidth]{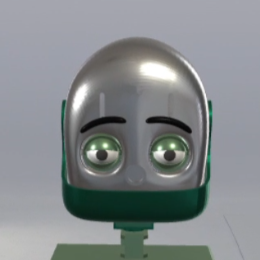}
         \caption{Sadness.}
         \label{fig:vr_face_sadness}
     \end{subfigure}
     \begin{subfigure}[t]{0.135\textwidth}
         \centering
         \includegraphics[width=\textwidth]{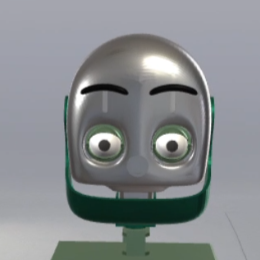}
         \caption{Surprise.}
         \label{fig:vr_face_surprise}
     \end{subfigure}
     \begin{subfigure}[t]{0.135\textwidth}
         \centering
         \includegraphics[width=\textwidth]{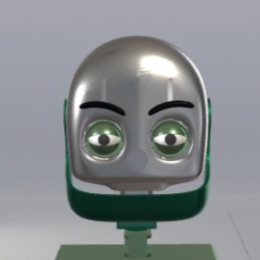}
         \caption{Disgust.}
         \label{fig:vr_face_disgust}
     \end{subfigure}
     \begin{subfigure}[t]{0.135\textwidth}
         \centering
         \includegraphics[width=\textwidth]{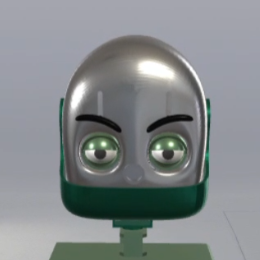}
         \caption{Anger.}
         \label{fig:vr_face_anger}
     \end{subfigure}
     \begin{subfigure}[t]{0.135\textwidth}
         \centering
         \includegraphics[width=\textwidth]{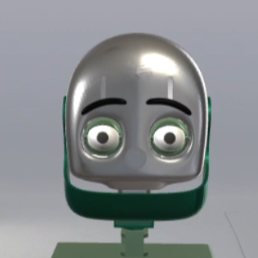}
         \caption{Fear.}
         \label{fig:vr_face_fear}
     \end{subfigure}
     \vspace{-2mm}
    \caption{Robot facial expressions.}
    \label{fig:expressions}
    \vspace{-6mm}
\end{figure*}

A robot must be able to make facial expressions and identify them in order to engage and communicate with people more effectively~\cite{liu2017}. This experiment shows how the robot may convey his emotions through his expressions. Their simplicity and quality are essential in the social HRI field, making this part vital to the framework.
Fig.~\ref{fig:expressions} shows the set of different facial expressions performed by the robotic face.

Additionally, the framework offers a preprogrammed feature of facial expression recognition. Through this, researchers can develop applications where, for example, the robot can mimic a human's mood and understand the context of the dialogue by analyzing human emotion. The recognizer system was tested with the platform 3D avatars and showed excellent accuracy in expressions of joy, sadness, surprise, neutral, and anger; meanwhile, it presented a poor performance with 3D avatars containing fear and disgust facial expressions.

\subsection{Manipulators}\label{manipulators}
In order to enrich the development possibilities of HRI applications, our platform offers not only an animatronic face but also a functional robotic body composed of a fixed base, a torso, and two manipulators as arms.
That structure allows the performance of body poses and gestures through fluid motions and more complex activities, like manipulation tasks.
This broadens the scope of applications that can be created utilizing the framework.

\begin{figure*}[bp!]
    \vspace{-2mm}
    \centering
    \includegraphics[width=\textwidth]{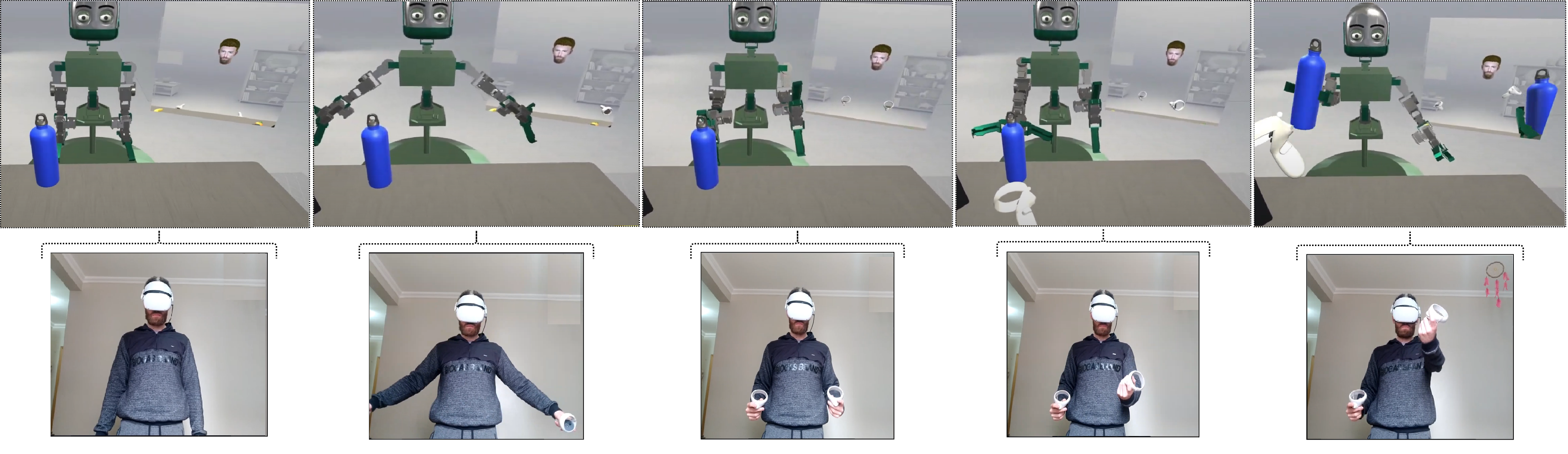}
    \vspace{-6mm}
    \caption{Human operating robot manipulators through VR controllers.}
    \label{fig:operation}
\end{figure*}

The robot can therefore interact with things in the VR world, in addition to interacting with people, as seen in Fig.~\ref{fig:operation}, which demonstrates the manipulator grabbing an object. This competency was created to make it easier for developers to construct and test movement patterns without using a real robot, speeding up the creation of any application that uses a similar manipulator with the same structure. In this experiment, an object was picked by the robot, lifted, and placed back on the table via a series of preprogrammed moves.

\subsection{Operating the Robot}\label{operating}
Still focusing on the simulated robot body capacities, the VR platform provides the capability for users to operate the manipulators and neck in real-time.
The operation of the simulated robot is made through VR controllers and a VR headset, in which each 6DOF controller operates a single manipulator, and the headset orientation controls the robot neck.
Fig.~\ref{fig:operation} shows that the user is capable of controlling the robot arms by doing natural human movements.

This approach allows, for example, applications of imitation, where the robot mimics the human gesture while interacting and enables the recording of preprogrammed fluid motions, which can be acquired through natural movements made by the user and applied in the simulated robot afterwards. 
Furthermore, the operation of the robot can be made from a first-person perspective, where the user has the possibility to control the robot joints from the robot's visual perspective. This technique can be implemented to teleoperate a real robot with the same physical structure~\cite{franzluebbers2019remote}.


\section{DISCUSSION}\label{discussion}

The experiments session demonstrated the effectiveness of the VR approach for executing and testing HRI applications.
As a platform made for researchers, numerous facilities have been created to aid the framework's user in more efficiently developing his study.
For example, users do not need to worry about minor code implementations because our framework offers a variety of easy-to-use and full-featured implemented algorithms.

As demonstrated in the IV-A subsection, our architecture already includes visual tracking for objects or people's faces. With the help of these tools, researchers can develop studies in behavioral analysis and attention-biased experiments, as well as perform a variety of tests with a focus on color-based visual tracking and the functionality of easy recognition, attention, and optical monitoring of an individual or group of people interacting socially with the robot. The neck and eye tracking make the immersive experience more humanized to the VR user.

As mentioned in IV-B subsection, the framework's built-in face recognition technology enables the researcher to engage in more precise interactions that increase the user's sense of immersion and reliability while interacting with the robot. Also, as demonstrated in IV-C subsection, the humanoid's face can produce different emotions to provide rich social interaction with humans. The ability to make and recognize facial expressions opens up a variety of possibilities for study into the psychological and behavioral underpinnings.

Our platform provides simple access to control the robotic manipulation system tested in the IV-D subsection. From there, it is possible to conduct experiments by grabbing things and trading them with the robot~\cite{mathur2022review}. Using image-based pick and place and grasping algorithms, the researcher can quickly enhance the framework manipulation system~\cite{yang2020}. To do that, the developer only needs to communicate the algorithm with the framework's established ROS nodes. The researcher will thus be free to investigate new manipulation approaches and subsequently create additional HRI applications.

Also, the fact that voice commands were employed throughout all the previous experiments as the primary mean of communication between the user and the humanoid robot encourages the implementation of our methodology in additional research fields.
In many contexts, the usage of educational robots in teaching sectors has already been shown to be effective~\cite{OSPENNIKOVA201518,anwar2019systematic}. For instance, this sort of application can be enhanced by using the platform suggested in this study to elaborate robotics and programming lessons and demonstrations that involve direct student-robot interaction. It could be achieved in future works by using a multiplayer VR feature that enables multiple users to access the same virtual world and interact with the robot and each other.

Furthermore, it is essential to emphasize that the platform is not limited to being used just for human-robot social interaction situations. 
As demonstrated in IV-E subsections, the framework offers an operating functionality that allows the user to control the joints of the robot. 
That broadens the possibility of exploration in studies involving manipulation tasks. For example, applying imitation learning after collecting demonstrations of humans controlling the robot manipulator to accomplish determined tasks~\cite{imitation_learning_vr}.

Finally, since the framework allows roboticists to test and evaluate their algorithms and experiments without the requirement of the actual robot, considerably speeding up the evaluation process, our platform makes it possible to test, for example, computer vision and reinforcement learning algorithms without physical hardware and actual environment. In this sense, the researcher can manage their methods and certify their results interactively with the VR environment through the framework.

\section{CONCLUSIONS}\label{conclusions}

In this work, it was presented an open-source robotics framework for a virtual reality setting that enables an immersive and realistic human-robot interaction experience, which is possible even without the actual robot's hardware. 
The platform provides an excellent tool for any research and development in the field of HRI since it has tremendous potential for testing and developing new functionalities for the robot while significantly lowering costs, robot damage risks, and evaluation test time.

The trials have shown the robot's aptitude for expressing his emotions, recognizing speech and facial expressions, visually tracking and interacting with people and objects, and being operated by the user.
The robot performed well in the experiments explored in this work, showing that researchers can use our framework's pre-built features to develop more complex simulated or even real-world applications.

Hence, the experiments demonstrate that a VR platform is a suitable and powerful tool in the construction of HRI applications, in which the robot can perform sociable interactions with humans, similar to a real situation.
In order to enhance the framework and increase the simulated robot's capabilities, there are many functionalities to be implemented, for example, object recognition, emotion extraction from human spoken words, a set of preprogrammed body gestures, and VR multiplayer feature.




\section*{Acknowledgement}

\anonymize{The authors would like to thank the VersusAI team.}

\bibliographystyle{./bibliography/IEEEtran}
\bibliography{./bibliography/IEEEabrv,./bibliography/IEEEreferences}

\addtolength{\textheight}{-12cm}  

\end{document}